\pdfoutput=1

\documentclass[11pt]{article}

\usepackage{acl}

\usepackage{times}
\usepackage{latexsym}
\usepackage{tabularx}

\usepackage[T1]{fontenc}

\usepackage[utf8]{inputenc}

\usepackage{microtype}
\usepackage{booktabs}
\usepackage{graphicx}
\usepackage{placeins}
\usepackage{makecell}
\usepackage{multirow}

\newcommand{\tydi}{{\sc TyDi QA}}

\title{GAAMA 2.0:\\ An Integrated System that Answers Boolean and Extractive Questions}

\author{Scott McCarley, Mihaela Bornea, Sara Rosenthal,
 \textbf{Anthony Ferritto}, \\ \textbf{Md Arafat Sultan}, \textbf{Avirup Sil}, \textbf{Radu Florian} \\
        IBM Research AI \\
        Yorktown Heights, NY \\
        jsmc@us.ibm.com}

\begin{document}
\maketitle

\begin{abstract}

Recent machine reading comprehension datasets include extractive and boolean questions but current approaches do not offer integrated support for answering both question types. We present a multilingual machine reading comprehension system and front-end demo that handles boolean questions by providing both a YES/NO answer and highlighting supporting evidence, and handles extractive questions by highlighting the answer in the passage. Our system, GAAMA 2.0, is ranked first on the \tydi{} leaderboard at the time of this writing. We contrast two different implementations of our approach. The first includes several independent stacks of transformers allowing easy deployment of each component. The second is a single stack of transformers utilizing adapters to reduce GPU memory footprint in a resource-constrained environment.



\end{abstract}

\section{Introduction}

Current machine reading comprehension (MRC) systems \cite{alberti2019bert,chakravarti-etal-2019-cfo, ferritto-etal-2020-multilingual} typically feature a single model targeted at supplying short extractive answer spans, but boolean questions demand non-extractive yes/no answers, as well as supporting evidence.
We demonstrate here a system that, given a question, predicts the expected answer type and
provides direct YES/NO answers with supporting evidence to boolean questions, or provides short answers to extractive questions.
\footnote{core capabilities of our system are available at \url{https://github.com/primeqa/}}
See examples of both question types in Figure \ref{fig:bool-explainable-examples}.

\begin{figure}[t]
    \centering
    \small
\framebox{%
  \begin{minipage}{\columnwidth}

1.Is the Mississippi the longest river in the world?  \textbf{NO} \\
\\
As a result, the length measurements of many rivers are only approximations (see also coastline paradox). In particular, \textbf{there seems to exist disagreement as to whether the Nile[3] or the Amazon[4] is the world's longest river.} The Nile has traditionally been considered longer, but in 2007 and 2008 some scientists claimed that the Amazon is longer[5][6][7] by measuring the river plus the adjacent Pará estuary and the longest connecting tidal canal.[8]
\\
\\ 
2. Roughly, how much oxygen makes up the Earth crust? \textbf{almost half of the crust's mass}
\\
Diatomic oxygen gas constitutes 20.8\% of the Earth's atmosphere. However, monitoring of atmospheric oxygen levels show a global downward trend, because of fossil-fuel burning. Oxygen is the most abundant element by mass in the Earth's crust as part of oxide compounds such as silicon dioxide, making up \textbf{almost half of the crust's mass.}

  \end{minipage}}    
  \caption{Examples of boolean and factoid questions.
  The factoid question requires an extractive answer; the boolean question requires a YES/NO answer and extracted supporting evidence
  }
    \label{fig:bool-explainable-examples}
\end{figure}
\begin{figure*}[t]
\includegraphics[width=\linewidth]{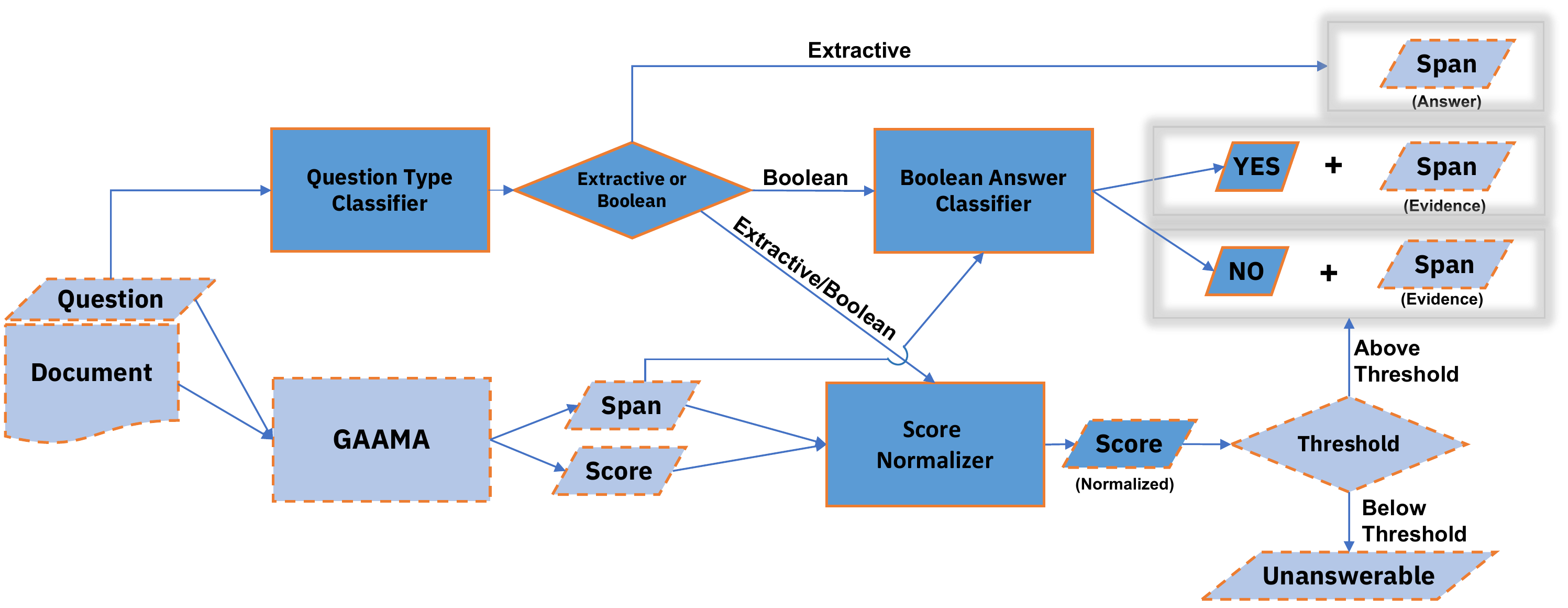}
\caption{System diagram: Pale blue boxes are the components of a traditional MRC system;  dark blue boxes are the additional components that are necessary for proper handling of both boolean and extractive questions.  See Section \ref{sec:systemDescription} for a full explanation.}
\label{fig:system_diagram}
\end{figure*}

\begin{figure*}[ht]
  \centering
  \includegraphics[width=\linewidth]{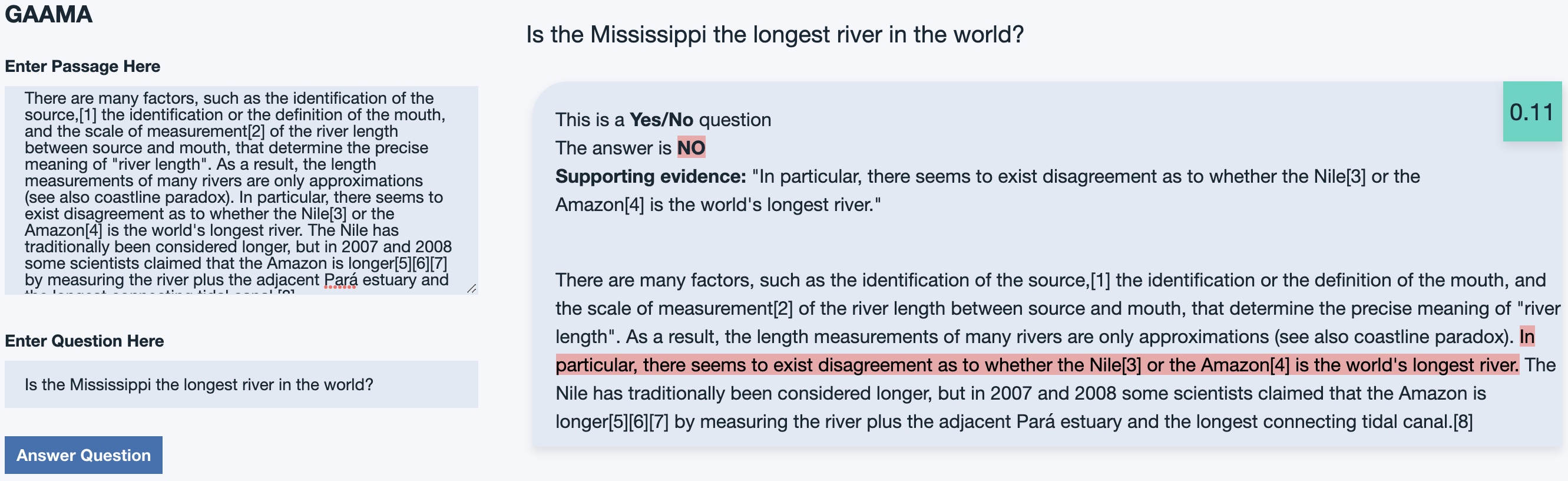}
   \caption{A screenshot of our demo.  The left side is user input.  On the right side, the system response identifies the question as a yes/no question, provides a succinct answer NO, and identifies the supporting evidence.}
   \label{fig:ui_demo}
\end{figure*}

We highlight several capabilities, beyond those of traditional extractive MRC, that are necessary for our demonstration:  the system must distinguish boolean and extractive questions,
must be able to generate a non-extractive YES/NO answer if the question is boolean, and must be able to recognize unanswerable questions regardless of whether it produces an extractive or non-extractive answer.



These new capabilities are needed because the development of MRC has been driven by training with extractive datasets \cite{squad, naturalQuestions}, while
boolean questions have been explored in isolation \cite{clark-etal-2019-boolq}.
Therefore,
extractive questions are typically handled by a pointer network which locates the start and end token of the answer span in the passage.
On the other hand, boolean questions are handled by a binary classifier that classifies an entire passage with a YES or NO answer, ignoring the need to also provide concise supporting evidence for the answer.
While these models are individually well-understood, we elucidate the design considerations that are necessary to present these capabilities to the user in an integrated manner.

The backend of the system is shown in Figure \ref{fig:system_diagram}.  
In addition to GAAMA (Go Ahead Ask Me Anything) \cite{chakravarti-etal-2019-cfo,chakravarti2020towards}, the core of our traditional MRC system, our system incorporates a {\em question type classifier} to distinguish boolean and extractive questions, a {\em boolean answer classifier} to provide YES/NO answers to the boolean questions, and a {\em score normalizer} component designed to improve the identification of unanswerable questions.

The user interface of our system is shown in Figure \ref{fig:ui_demo}.
When our system is presented with a boolean question, it indicates that it has identified the question as boolean.
It provides a direct YES/NO answer, and
provides supporting evidence, which is highlighted in the paragraph.



The contributions of this paper are:
\begin{enumerate}
    \item We provide an integrated system architecture and front-end demo that appropriately handles extractive \textit{and} boolean questions. We describe and analyze each component of our pipeline.
    \item We investigate parameter sharing via adapters \cite{pmlr-v97-houlsby19a} as a modeling choice to reduce the GPU memory footprint
    of the system in a resource-constrained environment.
    \item We present end-to-end MRC results on the \tydi{} \cite{clark-etal-2020-tydi}  
    dataset, which has a mixture of extractive and boolean questions. Our system backend achieves state-of-the-art results on the \tydi{} leaderboard.

\end{enumerate}

\section{Related Work}

Our demo is an extension of prior works GAAMA ~\cite{chakravarti-etal-2019-cfo} and M-GAAMA \cite{ferritto-etal-2020-multilingual}. Both are QA demos with the former being English\textit{-only} and the latter being cross-lingual. We add the capability to answer both extractive \textit{and} boolean questions in one integrated multilingual system, thus creating GAAMA 2.0.


Other QA demos include: ARES \cite{ferritto-etal-2020-ares}, a QA demo that features ensembling of several MRC systems, BERTSerini \cite{yang2019endtoend} , which leverages the Anserini IR toolkit \cite{Yang:2017:AEU:3077136.3080721} to extract relevant documents given a question in English only, NAMER \cite{zhang-etal-2021-namer} for multi-hop knowledge base QA, and Talk to Papers \cite{zhao-lee-2020-talk} for QA in academic search.


We make use of two datasets in developing our system: \tydi{} \cite{clark-etal-2020-tydi} and BoolQ-X \cite{rosenthal2021answers}. \tydi{} is a multilingual MRC dataset containing questions in multiple languages. 10\% of the questions are boolean. The \tydi{} boolean question annotations include the paragraph the answer is found in and a label: YES or NO. A boolean question can also be unanswerable. BoolQ-X is an enhancement of BoolQ \cite{clark-etal-2019-boolq}, a large dataset of over 18k boolean questions. BoolQ-X modifies the short passages of BoolQ, 
to make it more suitable for MRC by expanding the original answer passage while preserving its context in the original Wikipedia document. Other datasets that contain a mixture of boolean and extractive questions are Natural Questions \cite{naturalQuestions} and MS Marco \cite{DBLP:conf/nips/NguyenRSGTMD16}. These datasets are only English and contain a smaller percentage of boolean questions than \tydi{}.

\section{System Components}
\label{sec:systemDescription}


A diagram of the system is illustrated in Figure \ref{fig:system_diagram}.
The light blue boxes depict a traditional MRC system.  The question and passage are jointly passed to GAAMA 
which finds an answer span in the passage, and a score.  
If the score is below a threshold, the question is assumed to be unanswerable.
The additional components
required for our system are shown in the dark blue boxes in Figure \ref{fig:system_diagram}.
The {\em question type classifier} determines if the question is boolean or extractive.  
If the question is extractive, the span produced by GAAMA is assumed to be the extracted answer span.  
On the other hand, if the question is boolean, the question/passage pair are passed to a {\em boolean answer classifier} that decides if the passage supports an answer of YES or NO. 
The span produced by GAAMA is then treated as supporting evidence for the boolean answer.
The output of the question type classifier is also used by the {\em score normalizer}, which enables the thresholding mechanism to more reliably determine if the question is answerable.


\subsection{GAAMA}
\label{sec:gaama}
GAAMA is a single component that extracts a candidate answer span from the question/document pair. 
This component extends a traditional extractive question answering system \cite{ferritto-etal-2020-multilingual}, and is implemented with a pointer network head
on the 24 layer xlm-roberta-large \cite{conneau-etal-2020-unsupervised}. 
We use multi-teacher knowledge distillation to distill the knowledge of both a \tydi{} model and a Natural Questions model into a single robust student model, using all training examples from the two datasets.
Synthetic training data was also used in this model \cite{chen2020improved}.
Our model is trained with the short answer for extractive questions, and the passage as a span for boolean questions, as the 
\tydi{} training data does not provide a short answer for boolean questions.
At runtime, this component is agnostic to the difference between boolean 
and extractive questions, producing only a span of extracted text.


\subsection{Question Type classifier}

The question type classifier takes as input the question, and returns a label that distinguishes boolean and extractive questions.
It is a  multi-lingual transformer-based (MBERT \cite{devlin-etal-2019-bert})  classifier.
The question type classifier was trained and evaluated on the answerable subset of the \tydi{} questions.
Even though there are boolean questions that are unanswerable, we cannot use these questions to train or evaluate our model because \tydi{} did not provide labels to indicate whether the unanswerable questions were boolean.
The classifier achieved an F1 score of $99.2\%$ on boolean questions, and $94.6\%$ on factoid questions for the \tydi{} dev set.
Our model achieves high F1 score for all \tydi{} languages.
Because this accuracy was extremely high, we did not pursue further refinements.
We believe this task performs well because there are certain words that signal with high likelihood that a question is boolean or not. For example, in English, questions starting with \textit{is}, \textit{does}, and \textit{are}  are usually boolean questions while the 5 W questions (\textit{who}, \textit{what}, \textit{why}, \textit{where}, \textit{when}) indicate extractive questions.

\subsection{Boolean Answer Classifier}

The boolean answer classifier is a binary classifier that predicts
a YES or NO answer to the question,
given a boolean question and a passage. 
This component is only invoked if the question type classifier has
determined that the question is boolean.

We trained the classifier using upstream system output: boolean questions from \tydi{} data, as selected by our question type classifier, along with the corresponding system output passages 
selected by the GAAMA component.
There are often multiple passages containing the correct answer and GAAMA may find one not in gold. This also mimics what occurs during real-world use of this component.  
In addition, we supplemented the \tydi{} training data with questions and 200 word passages
selected from BoolQ-X \cite{rosenthal2021answers}, which is more compatible with the MRC task and the \tydi{} data than BoolQ.

\begin{figure}[t]
  \centering
  \includegraphics[width=1\linewidth]{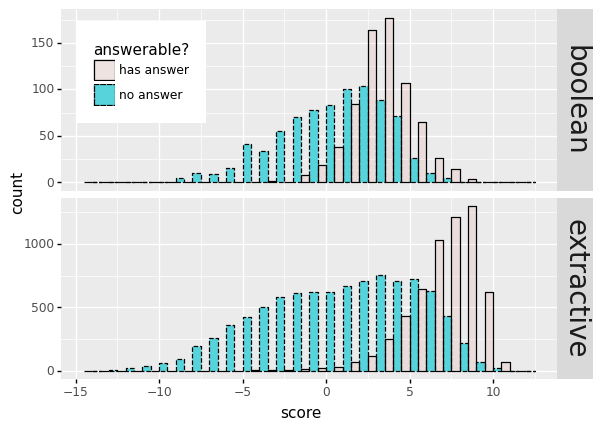}
   \caption{Score distributions of answerable and non-answerable questions differ between boolean and extractive questions.}
   \label{fig:score_histogram}
\end{figure}


A simple baseline for the boolean answer classifier is to answer all boolean questions with YES. This is a particularly strong baseline for \tydi{} since the boolean data is so skewed with 80\% of answers being YES. This majority baseline has a YES/NO F1 of 90.1/0.0 for \tydi{} and 77.0/0.0 for BoolQ-X. While the F1 for YES is high for both datasets, the F1 for NO is 0.0. In addition a system that always answers YES would of course not be satisfying to a user. Our boolean answer classifier obtains a YES/NO F1 of 91.0/44.5  on the \tydi{} dev set and 81.3/65.7 on the BoolQ-X dev set. Our YES F1 is similar to the \tydi{} baseline and we achieve a large improvement for NO. Further, on the more balanced BoolQ-X dataset, we achieve a large improvement for both classes. We remind the reader that BoolQ-X is designed to reflect more realistic conditions that are compatible with the MRC task and should not be compared with scores from the original BoolQ.

In addition to training the classifier with the full passage, we also experimented with classifying the minimal answer span produced by GAAMA that is used as evidence. We found that it is preferable to expand the context seen by the classifier beyond the spans produced by the upstream MRC system, rather than using the MRC system output spans as-is. Note that this is a very data-specific heuristic. While evaluating \tydi{} we use the paragraph boundaries provided in the dataset, but in the demo system we simply expand the system-generated spans into wider windows based on fixed offsets surrounding the begin and end of the spans.

\subsection{Score normalizer}
\label{sec:normalizer}

The score normalizer scales the score produced by GAAMA for the answer span to the $[0,1]$ interval, conditioned on  question type.
This score is used to determine whether the question is answerable or not. The initial distribution of scores from GAAMA
are strikingly different depending upon whether the question is boolean or extractive.
This effect is illustrated in Figure \ref{fig:score_histogram}. The answer spans in the \tydi{} dataset are typically a few words. However, since the boolean questions do not have answer spans, we supply the candidate paragraph as the evidence span for supervision during training. This mismatch in span length, and the infrequency of boolean questions causes the scores of boolean questions to be assigned  a low confidence, often below the threshold making the majority of boolean questions unanswerable. This threshold is determined by the official \tydi{} evaluation script. 




\begin{table}[t]
\centering


\begin{minipage}{\linewidth}
\small
    \begin{tabular}{c|c||c|c||c|c}
     \multicolumn{2}{c||}{} & \multicolumn{2}{c||}{Boolean (YN)} &
     \multicolumn{2}{c}{Extractive (MA)}\\
     \midrule
     \multicolumn{2}{c||}{Threshold}& Above & Below &  Above &  Below \\
    \toprule
    \multirow{3}{*}{\rotatebox[origin=c]{90}{w/o SN}} & YN & 259 & \textbf{826} & 14 & 16 \\
     & MA & 31 & 37 & 5,590 & 1,195\\
     & NA & 22 & 776 & 1,430 & 8,137\\
    \toprule
    \multirow{3}{*}{\rotatebox[origin=c]{90}{w/ SN}} & YN & \textbf{781} & 304 & 13 & 17 \\
     & MA & 56 & 12 & 5,432 & 1,353\\
     & NA & 128 & 670 & 1,230 & 8,673\\
     \bottomrule
    \end{tabular}
    \caption{Confusion matrices with and without the score normalizer (SN) on \tydi{} DEV. The columns are the predictions: Y/N and extractive answers that score above or below the threshold. Predictions below the threshold are marked as unanswerable. The rows are the gold labels: YN for the boolean questions, MA for the questions with a minimal (extractive) answer, and NA are the questions with no answer.}
    \label{tab:confusion-sn}
\end{minipage}
\end{table}

Table~\ref{tab:confusion-sn} shows the confusion matrix for our system with and without using score normalization. The question type classifier identifies the majority of boolean questions correctly, however the majority of the gold YN questions (\textbf{826}) are \textit{below} the threshold as shown in the first line of the table. After the application of the score normalizer, the bulk (\textbf{781}) of the gold YN questions are now above the threshold in the fourth line of Table~\ref{tab:confusion-sn}. Meanwhile, the score normalizer did not have a significant impact on the gold minimal answer (MA) for extractive questions, that already had the correct predictions, only 2\% moved from above to below the threshold. Further, we increased the number of questions correctly identified as unanswerable, 2\% of the extractive NA questions moved from above to below threshold.


The score normalizer is implemented as a logistic regression classifier using the output
of the question type classifier and the span score of the GAAMA system as features.
It generates a probability of whether the question/passage pair is marked as \textit{answerable} or \textit{unanswerable}.  We used the \tydi{} annotations to find the labels for training the score normalizer.  For Table \ref{tab:confusion-sn} we built a score normalizer from a held out 10\% of the \tydi{} train.
The new score is the probability of the \textit{answerable} class. 

\FloatBarrier

\section{End-to-End MRC Results}


In this section we present the end-to-end results of our system,
as measured on \tydi{}.
 We divide the  \tydi{} dev in  two parts. The first half was used for calibrating the score normalizer for the end-to-end system and the second half was used for validating our model during development. We validate our approach on the official test set used in the \tydi{} leaderboard via blind submission.
All experiments were implemented using the Huggingface frameworks from \cite{wolf-etal-2020-transformers}.



\begin{table}[bt!]
\small
\centering
\begin{tabular}{c|c|c}
\toprule
&\textbf{System} & \textbf{MA} \\
\toprule

\multirow{2}{*}{\rotatebox[origin=c]{90}{DEV}}&\hbox{GAAMA}           & 68.6 \\
    &\hbox{GAAMA 2.0} & \textbf{72.6 }\\
\midrule
\multirow{3}{*}{\rotatebox[origin=c]{90}{TEST}}&\hbox{GAAMA 2.0}   & \textbf{72.3} \\
    &\hbox{GAAMA-DM-Syn-ARES}&  68.0  \\
    &\hbox{PoolingFormer }  & 67.6  \\
\bottomrule
\end{tabular}
\caption{End-to-End results on half of the \tydi{} dev set and \tydi{} test set.  Test set results are evaluated by the \tydi{} leaderboard. We show the F1 score on the Minimal Answer (MA) prediction task.
Our GAAMA 2.0 \tydi{} leaderboard submission was called GAAMA-Syn-Bool-Single-Model.}

\label{tab:tydi-results}
\end{table}

In Table \ref{tab:tydi-results}, we show minimal answer (MA) \tydi{} results on the second half of the \tydi{} dev set.
Our baseline system, GAAMA, is an xlm-roberta model trained on the \tydi{} data as described in Section \ref{sec:gaama}. This system does not handle boolean questions. The GAMMA 2.0 system implements the architecture in Figure \ref{fig:system_diagram}, including the \textit{question type classifier}, the \textit{boolean answer classifier} and the \textit{score normalizer} components. Our experiments show a clear gain of 4 F1 points which is credited to the correct prediction of boolean questions. It is worth noting that our pipeline does not hurt the performance of extractive questions. Based on the analysis in Table \ref{tab:confusion-sn}, only 0.01\% of the extractive (MA) gold questions that have an answer are incorrectly classified as boolean. An additional 0.01\% of the gold questions that have no answer (NA) in \tydi{} have an incorrect boolean prediction. We showed in Section~\ref{sec:normalizer} that we don't affect the performance of extractive questions and our gains are purely on handling boolean questions.

\begin{table}[t]
\small
\centering
    \begin{tabularx}{\columnwidth}{|X|X|X|X|X}
\toprule
\textbf{GAAMA configs} & \textbf{F1} & \textbf{\# params $(\times 10^6)$} & \textbf{size (MiB)} \\
\midrule
\textit{Separate} & 72.6 & 1680 & 3204 \\
\textit{Adapters} & 73.0 & 563 & 1074 \\
\bottomrule
\end{tabularx}
\caption{A comparison of GAMMA 2.0 using separate models and adapters. The F1 score is the minimal answer reported by running the end-to-end system on the \tydi{} dev set.}
\label{tab:footprint}
\end{table}

A blind submission of our system, GAAMA-Syn-Bool-Single-Model (GAAMA 2.0) to the  \tydi{} leaderboard\footnote{\url{https://ai.google.com/research/tydiqa, as of 2022-01-12}} achieved state of the art performance on the hidden test set. The \tydi{} leaderboard has attracted research targeting different aspects of  the document level, multilingual question answering task. To the best of our knowledge GAAMA 2.0 is the only \tydi{} submission that handles boolean questions. There is a notable improvement of over 4 F1 points compared to the next best system, GAAMA-DM-Syn-ARES, and almost 5 F1 points compared to the third place system, PoolingFormer. GAAMA-DM-Syn-ARES is a variation of the GAAMA system featuring an ensemble \cite{ferritto-etal-2020-ares} of systems targeting representations with long document dependencies \cite{document_modeling}. These features are not included in the GAAMA baseline we use. PoolingFormer \cite{zhang2021poolingformer} also focussed on modeling long dependencies in documents and they modify the self attention mechanism in the transformers to enable long input sequences. It is worth noting that our system can work with these and other MRC approaches.










\section{Parameter sharing approach}

In the baseline implementation of our system, each of the transformer-based classifiers is a separate stack of transformers, fine-tuned independently for its particular task\footnote{Although a single joint transformer model for all components of our model is conceivable, we do not explore this approach because the capabilities of the components are developed independently, and the cost of repeated retrainings of a joint model during development would be prohibitive.}.
This implementation is convenient because the components can be developed and deployed independently. Each of the classifiers can be deployed in a microservice in a separate docker container. The containers may be deployed on either the same machine or different machines, depending upon the availability of GPUs. The communication between the microservices can be easily handled by the flow compiler of \cite{chakravarti-etal-2019-cfo}. On the other hand, deploying multiple transformer-based classifiers is expensive, since they must remain resident in GPU memory.  In practice this requires separate GPUs for each of our three transformer-based classifiers. 

To address this concern, we also experiment with adapter-based models.
With adapters \cite{Houlsby2019ParameterEfficientTL}, multiple stacks of transformers can be replaced with a single stack of transformers, and a set of {\em adapters}, which have $<1\%$ of the number of parameters of 
the corresponding transformer stack.
Only the parameters in the adapter need to be fine-tuned for a particular task - the parameters of the transformer stack itself are shared between the multiple tasks. 
Adapters have been used successfully for many classification tasks.

Adapters are typically applied to the shared parameters of a transformer stack pretrained solely for masked language modeling \cite{pfeiffer-etal-2020-mad, pfeiffer-etal-2021-adapterfusion}.
In contrast, we view our span extraction transformer as the base task, and train the other tasks as adapters on top of the span extraction transformer stack. This allows the other components to leverage the MRC data our span extraction model was trained on. Further, training our span extractor is a specialized process involving distillation from multiple teacher models, and is not easily amenable to the adapter framework.



We implement the query type classifier 
and the boolean answer classifier with adapters inserted into our span extractor model, using the framework of \cite{pfeiffer2020AdapterHub}.  The adapter-based question type classifier achieves an F1 score of $99.6\%$ on boolean questions, and $97.3\%$ on extractive questions on the \tydi{} dev set. This is notably better than the original implementation. The adapter-based boolean answer classifier also achieved comparable or better performance to the original implementation with a YES/NO
F1 score of 90.9/41.5\footnote{The drop in F1 of NO in the \tydi set is likely negligible because of the small size and the extreme skew.} for \tydi{} and 83.0/68.8 for BoolQ-X.
We compare end-to-end MRC performance of both approaches in Table~\ref{tab:footprint}. The F1 scores are similar for both approaches, while the adapter system reduces our memory footprint significantly.

\section{Conclusion}

We present a demonstration of a machine reading comprehension system that can answer both boolean questions and factoid questions in an integrated system. When a question is boolean, it provides a direct YES/NO answer and highlights the supporting text. When a question is extractive it highlights the answer span found in the text. These new  capabilities require adding additional components to a traditional MRC system:  a {\em question type classifier},  a {\em boolean answer classifier} and a {\em score normalizer}. Each component is an important part of the design needed to successfully answer boolean questions. Our back-end system achieves a four point improvement over the comparable system without boolean questions and achieves state-of-the-art results on the \tydi{} leaderboard.

Finally, we contrast the merits of two different implementation approaches. In one, we implement each of the components in a separate microservice for flexibility.  In the other, we apply a single transformer via adapters to reduce the GPU memory footprint and associated expense. The adapters also enabled the question type classifier and boolean answer classifer components to leverage additional training data not directly useful to the boolean questions. In the future we would like to explore incorporating other question and answer types into our demo, such as how-to questions that require a list as answer.

\FloatBarrier

\bibliography{acl_bool}
\bibliographystyle{acl_natbib}


\end{document}